\documentclass{article}

\usepackage{makecell}
\usepackage[final]{lg}

\usepackage[utf8]{inputenc} 
\usepackage[T1]{fontenc}    
\usepackage{hyperref}       
\usepackage{url}            
\usepackage{booktabs}       
\usepackage{rotating}
\usepackage{amsfonts}       
\usepackage{amsmath}
\usepackage{nicefrac}       
\usepackage{microtype}      
\usepackage[dvipsnames]{xcolor}
\usepackage{kotex}
\usepackage{adjustbox}
\usepackage{multirow}
\usepackage{amssymb}
\usepackage{longtable}
\usepackage{comment}
\usepackage[autostyle]{csquotes}
\usepackage{hhline}
\usepackage{tabu}
\usepackage{tablefootnote}
\usepackage{longtable}
\usepackage{graphicx}
\usepackage{array}
\usepackage{caption}
\usepackage{tcolorbox}
\usepackage{lipsum}
\usepackage{relsize}
\usepackage{colortbl}
\usepackage{tablefootnote}
\tcbuselibrary{breakable}
\usepackage{floatpag}
\floatpagestyle{plain}
\usepackage{float}
\usepackage{threeparttable}

\usepackage{color}
\usepackage{soul}
\definecolor{lightlightgray}{rgb}{0.9,0.9,0.9}

\usepackage{ulem}
\newcommand\kohlgrey[1][lightlightgray]{%
  \bgroup 
  \markoverwith{\textcolor{#1}{\vrule width.1em height.8em depth.2em}}%
  \ULon 
}

\usepackage{pifont}

\newcommand{\model}{EXAONE Deep }
\newcommand{\comp}{LG~AI~Research}

\title{\centering EXAONE Deep: Reasoning Enhanced Language Models}
\author{%
  \comp\thanks{The complete list of authors who contributed to this work can be found in Appendix~\ref{appendix:contributors}.
}\\
}

\begin{document}

\maketitle
\addtocounter{footnote}{-1}

\begin{abstract}

We present \model series, which exhibits superior capabilities in various reasoning tasks, including math and coding benchmarks. We train our models mainly on the reasoning-specialized dataset that incorporates long streams of thought processes. Evaluation results show that our smaller models, \model 2.4B and 7.8B, outperform other models of comparable size, while the largest model, \model 32B, demonstrates competitive performance against leading open-weight models. All \model models are openly available for research purposes and can be downloaded from \url{https://huggingface.co/LGAI-EXAONE}.

\end{abstract}


\begin{figure}[h!]
    \centering
    \includegraphics[width=\textwidth]{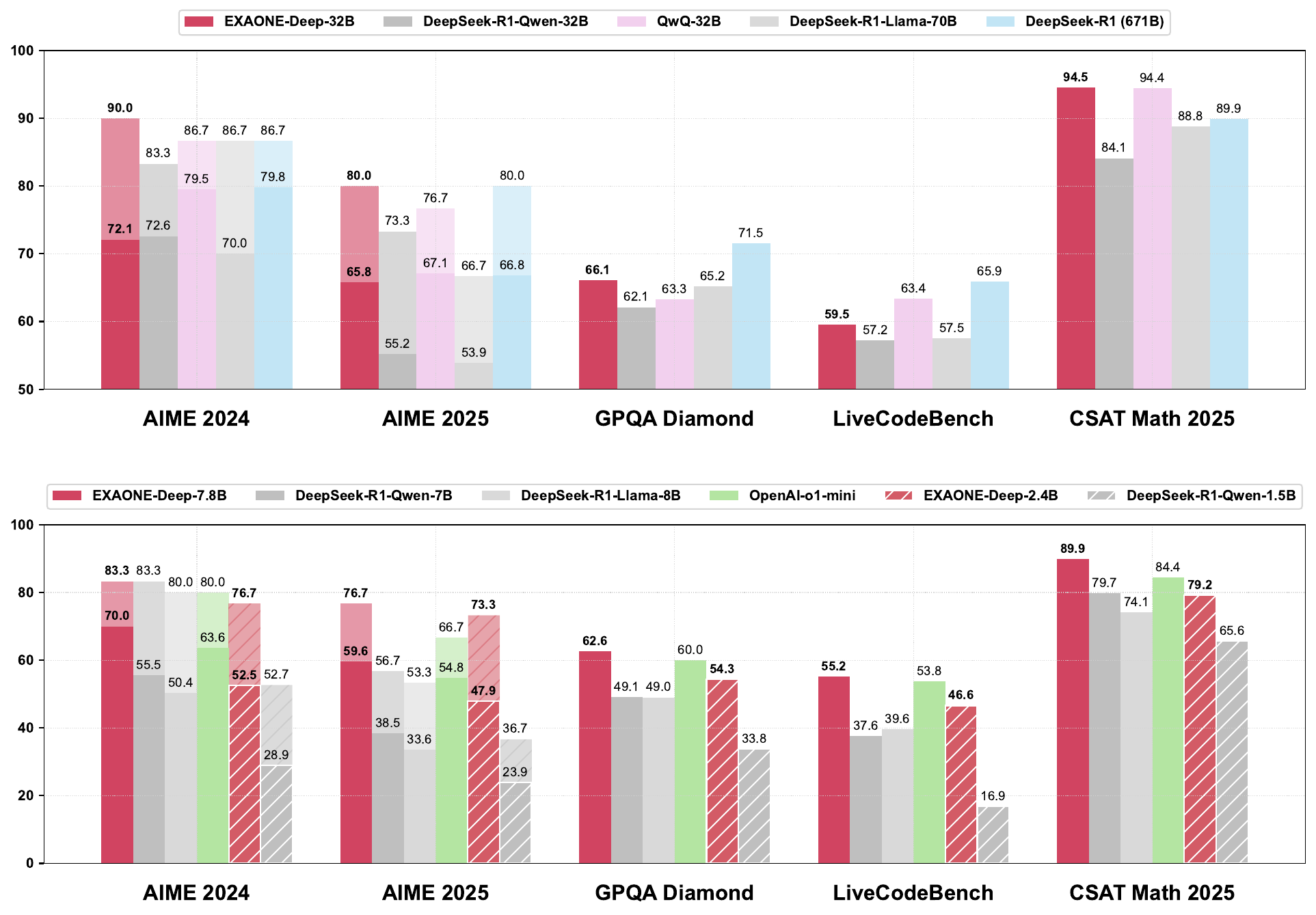}
    \caption{Overall performance comparison. EXAONE Deep 32B model demonstrates competitive performance compared to leading open-weight reasoning models such as QwQ-32B and DeepSeek-R1. It also outperforms both DeepSeek-R1-Distill-Qwen-32B and DeepSeek-R1-Distill-Llama-70B. The lightly colored regions in AIME 2024 and 2025 show the performance of  majority vote (consensus).}
    \label{fig:overall_performance_comparison}
\end{figure}

\newpage

\section{Introduction}

Recently, there has been a growing trend in research to enhance reasoning performance by adjusting computing resources during the testing phase \citep{snell2024scalingllmtesttimecompute}. In response to this trend, LG AI Research is introducing a new model lineup called \model 2.4B, 7.8B, and 32B. These models are fine-tuned versions of the EXAONE 3.5 series \citep{research2024exaone35serieslarge}, specifically optimized for reasoning tasks. We have trained these models using three prominent techniques widely employed for fine-tuning: Supervised Fine-Tuning (SFT), Direct Preference Optimization (DPO), and Online Reinforcement Learning (Online RL). 

The performance evaluation of the models indicates that, for the 2.4B variant, it demonstrates superior performance compared to DeepSeek-R1-Distill-Qwen-1.5B \citep{deepseekai2025deepseekr1incentivizingreasoningcapability}. For the 7.8B variant, it outperforms not only open-weight models of comparable scale such as DeepSeek-R1-Distill-Qwen-7B and DeepSeek-R1-Distill-Llama-8B \citep{deepseekai2025deepseekr1incentivizingreasoningcapability} but also a proprietary reasoning model OpenAI o1-mini \citep{openai2024o1systemcard}. In the case of the 32B model, its performance is competitive to that of the leading open-weight reasoning models such as QwQ-32B \citep{qwen2025qwq32b} and DeepSeek-R1 \citep{deepseekai2025deepseekr1incentivizingreasoningcapability}, and is superior to that of DeepSeek-R1-Distill-Qwen-32B and DeepSeek-R1-Distill-Llama-70B \citep{deepseekai2025deepseekr1incentivizingreasoningcapability} as illustrated in Figure~\ref{fig:overall_performance_comparison}.

\vspace{15pt}
\begin{center}
\noindent
\begin{minipage}{0.80\textwidth}  
    \textit{Thoughts become words, words become deeds, deeds become habits, habits become character, and character becomes destiny. Therefore watch the \textbf{<thought>}s of your mind with the sleepless eye of your mind.} \\
    \begin{flushright}
        \hfill \textsc{- Tryon Edwards}
    \end{flushright}
\end{minipage}
\end{center}
\section{Modeling}

\subsection{Data}

\begin{figure}[ht!]
    \centering
    \includegraphics[width=\textwidth]{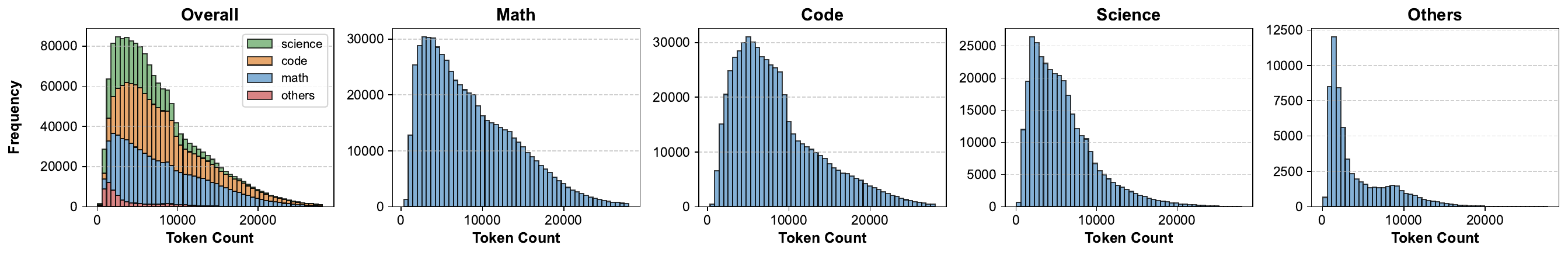}
    \caption{Distribution of token counts of the SFT dataset. Data points in the \texttt{Code} domain are notably longer on average, whereas those in the \texttt{Others} domain tend to be shorter.}
    \label{fig:token_length_distribution}
\end{figure}

To enhance the reasoning capabilities of language models, we have utilized 1.6M instances for SFT, 20K instances of preference data for DPO, and an additional 10K instances for Online RL. The SFT dataset contains approximately 12B tokens, with its length distribution illustrated in Figure~\ref{fig:token_length_distribution}. The dataset is designed to guide models in performing reasoning through an extended chain-of-thought (CoT) process, as illustrated in Figure~\ref{fig:sft_example}.

\begin{figure}[ht!]
\centering
\begin{tcolorbox}[
  title=SFT Dataset Example, 
  colframe=Black!80!White,   
  colback=gray!10,          
  coltitle=white,           
  colbacktitle=Black!80!White, 
  fonttitle=\bfseries,      
  breakable=false,          
  rounded corners,          
  width=\textwidth          
]

Query: Given the complex numbers $u$, $v$, and $w$ such that $uv = 12 + 16i$, $vw = 20 - 15i$, and $wu = -30 + 40i$, find real numbers $p$ and $q$ such that $u + v + w = p + qi$. Determine the value of $p^2 + q^2$.\\
\\
Response:\\
<thought>\\
Okay, so I need to find real numbers $p$ and $q$ such that $u + v + w = p + qi$, and then compute $p^2 + q^2$. The given information is the products of the complex numbers: $uv = 12 + 16i$, $vw = 20 - 15i$, and $wu = -30 + 40i$. Hmm, complex numbers can sometimes be tricky because of their real and imaginary parts, but maybe I can find expressions for $u$, $v$, $w$ individually?\\
\\
Let me think. If I have three equations involving the products of pairs of these complex numbers, maybe I can multiply all three equations together? Let's see: $uv \times vw \times wu = (uvw)^2$. So $(uvw)^2$ = $(12 + 16i)(20 - 15i)(-30 + 40i)$. If I compute that product, I can then take the square root to find $uvw$. But taking the square root of a complex number might give multiple values, but since we're dealing with complex numbers $u$, $v$, $w$, maybe we can choose a principal root or something. Wait, but maybe there's a better way.\\
\\
Alternatively, if I can express each variable in terms of another. For example, from $uv = 12 +16i$, maybe $u = (12 +16i)/v$. Similarly, from $vw = 20 -15i$, $w = (20 -15i)/v$. Then from $wu = -30 +40i$, substituting the expressions for $u$ and $w$ in terms of $v$, we get $[(20 -15i)/v] \times [(12 +16i)/v] = -30 +40i$. That simplifies to $(20 -15i)(12 +16i)/(v^2) = -30 +40i$. Then solving for $v^2$, we get $v^2 = (20 -15i)(12 +16i)/(-30 +40i)$. Then once we find $v^2$, we can take square roots to find $v$, and subsequently find $u$ and $w$. \\ ... \ (thought process continues) \\
</thought> \\ \\
For each $v$, we compute $u$ and $w$:\\
\\
Case 1: $v = \frac{9 - 13i}{5}$
\begin{itemize}
    \item $u = -2 + 6i$
    \item $w = \frac{15 + 5i}{2}$
\end{itemize}
Case 2: $v = \frac{-9 + 13i}{5}$
\begin{itemize}
    \item $u = 2 - 6i$
    \item $w = \frac{-15 - 5i}{2}$
\end{itemize}
Summing $u$, $v$, and $w$ in both cases:\\
\\
Case 1: $u + v + w = \frac{73}{10} + \frac{59}{10}i$\\
Case 2: $u + v + w = -\frac{73}{10} - \frac{59}{10}i$\\
\\
In both cases, $p^2 + q^2$ is:\\
\[
\left(\frac{73}{10}\right)^2 + \left(\frac{59}{10}\right)^2 = \frac{73^2 + 59^2}{100} = \frac{5329 + 3481}{100} = \frac{8810}{100} = \frac{881}{10}
\]\\
\\
Thus, the value of $p^2 + q^2$ is:\\
\[
\boxed{\dfrac{881}{10}}
\]\\

\end{tcolorbox}
\caption{An example of SFT dataset. The dataset is specifically designed to facilitate models in conducting reasoning tasks through an extended chain-of-thought methodology.}
\label{fig:sft_example}
\end{figure}

\subsection{Training}

The base models of \model are EXAONE 3.5 Instruct models \citep{lgai2004exaone3524binstruct, lgai2004exaone3578binstruct, lgai2004exaone3532binstruct}, which are instruction-tuned models possessing instruction-following capabilities. To enhance the reasoning abilities of EXAONE Deep, we structured SFT and DPO data in a templated format as illustrated in Figure~\ref{fig:sft_example}. Each training instance consists of a structured thought process followed by a final answer that synthesizes the reasoning steps into a coherent and precise response. Specifically, the EXAONE 3.5 Instruct models are trained to engage in reasoning within the \texttt{<thought>} and \texttt{</thought>} tags, performing step-by-step logical progression along with reflection, self-checking, and correction. The final answer, generated after reasoning, is designed to be self-contained, summarizing the key insights derived from the thought process in a clear and concise manner. This structured approach enables \model to engage in robust reasoning and deliver well-founded answers to a given query. We utilize SimPER \citep{xiao2025simperminimalistapproachpreference} as the training algorithm for DPO and our designed GRPO \citep{shao2024deepseekmathpushinglimitsmathematical} variant for Online RL.

In terms of training compute, the \model models are trained using NVIDIA H100 GPU clusters provided by Google Cloud Platform and NVIDIA NeMo Framework. The amount of computation used for pretraining of base models and fine-tuning of enhancing reasoning is presented in Table~\ref{tab:training_flops}.

\begin{table}[h!]
    \centering
    \setlength{\doublerulesep}{1pt}
    \begin{tabular}{wc{0.15\linewidth}|wc{0.15\linewidth}wc{0.15\linewidth}wc{0.15\linewidth}}
         \toprule
         Model size & Pretraining & Fine-tuning & Total \\
         \midrule
         32B & $1.25 \times 10^{24}$ & $7.04 \times 10^{21}$ & $1.26 \times 10^{24}$\\
         7.8B & $4.21 \times 10^{23}$ & $1.71 \times 10^{21}$ & $4.23 \times 10^{23}$ \\
         2.4B & $9.36 \times 10^{22}$ & $5.27 \times 10^{20}$ & $9.41 \times 10^{22}$ \\
         \bottomrule
    \end{tabular}
    \vspace{2mm}    
    \caption{Amount of computation (FLOPs) for model training.}
    \label{tab:training_flops}
\end{table}

\section{Evaluation}

\subsection{Benchmarks}

We evaluate models on MATH-500 \citep{lightman2023letsverifystepstep}, American Invitational Mathematics Examination (AIME) 2024 and 2025 \citep{maa2025aime}, the mathematics sections of South Korea's College Scholastic Ability Test (CSAT) 2025 \citep{kice2025suneung}, GPQA Diamond \citep{rein2023gpqagraduatelevelgoogleproofqa}, LiveCodeBench (24.08-25.02) \citep{jain2024livecodebenchholisticcontaminationfree}, MMLU \citep{hendrycks2021measuringmassivemultitasklanguage}, and MMLU-Pro \citep{wang2024mmluprorobustchallengingmultitask}.

In the case of CSAT, which includes textual problems and supplementary graphical information, the graphical content is excluded when evaluating the performance. Since most of this graphical information is also available in textual descriptions, its impact on results is considered minimal. In addition, students can choose from three elective subjects --- calculus, statistics, and geometry --- in the mathematics section. The final score is calculated as the average of the scores obtained in these three elective subjects.

\subsection{Baselines}

We have conducted comprehensive evaluations against strong baseline such as DeepSeek-R1 as well as comparable-scale baselines such as QwQ-32B, DeepSeek-R1-Distill-Qwen-32B, 7B, 1.5B, DeepSeek-R1-Distill-Llama-70B, 8B, and OpenAI o1-mini (2024-09-12).

\subsection{Evaluation Setup}

Following the setup described in the DeepSeek-R1 technical report \citep{deepseekai2025deepseekr1incentivizingreasoningcapability}, the maximum length of model generations is set to 32K tokens. Additionally, since the lengthy outputs of the reasoning models can vary greatly, as noted in the report, we adopt the pass@$k$ metric \citep{chen2021evaluatinglargelanguagemodels} to ensure the reliability of model performances. Specifically, we generate $k$ responses for each test case using a sampling temperature of 0.6 and a top-$p$ value of 0.95\footnote{For QwQ-32B, we additionally applied the recommended \(\text{top-} k = 40\) during evaluation. For OpenAI o1-mini, we were unable to adjust sampling parameters to our desired settings; thus, we used temperature = 1 and \( \text{top-} p = 1 \). }, and calculate pass@1 as followed:

\begin{equation} \label{equ:pass_at_1}
    \text{pass@1} = \frac{1}{k} \sum_{i=1}^{k}{p_i},
\end{equation}

where $p_i$ denotes the correctness of the $i$-th response. We also report cons@$k$~\citep{deepseekai2025deepseekr1incentivizingreasoningcapability} where the model generates $k$ answers and the final answer is the one generated the most frequently \citep{wang2023selfconsistencyimproveschainthought}.

The evaluation prompts used for assessing \model models are shown in Figure~\ref{fig:math_prompt_ours}, \ref{fig:mcqa_prompt_ours}, and~\ref{fig:code_prompt_ours}. For the CSAT benchmark, both prompts are employed, as it includes short-answer and multiple-choice questions. For baseline models, we report scores from official reports if available; otherwise, we measure performance ourselves. Specifically, for multiple-choice question answering types, we adopt prompts from OpenAI simple-evals framework\footnote{https://github.com/openai/simple-evals}. When evaluating short-answer questions, we utilize the MATH-500 prompt from simple-evals for OpenAI o-series models, but use the same prompt with ours (Figure~\ref{fig:math_prompt_ours}) for others, as it aligns with their recommended prompt.

\begin{figure}[ht!]
\centering
\begin{tcolorbox}[
  title=Prompt for Short-Answer Questions, 
  colframe=Black!80!White,   
  colback=gray!10,          
  coltitle=white,           
  colbacktitle=Black!80!White, 
  fonttitle=\bfseries,      
  breakable=false,          
  rounded corners,          
  boxsep=3pt,               
  width=\textwidth          
]

\{\{\textsl{question}\}\} \\
\\
Please reason step by step, and put your final answer within \textbackslash boxed\{\}.

\end{tcolorbox}
\caption{Prompt for evaluating models on short-answer questions. We apply the prompt to the MATH-500, AIME 2024/2025, and CSAT 2025 benchmarks.}
\label{fig:math_prompt_ours}
\end{figure}

\begin{figure}[ht!]
\centering
\begin{tcolorbox}[
  title=Prompt for Multiple-Choice Questions, 
  colframe=Black!80!White,   
  colback=gray!10,          
  coltitle=white,           
  colbacktitle=Black!80!White, 
  fonttitle=\bfseries,      
  breakable=false,          
  rounded corners,          
  boxsep=3pt,               
  width=\textwidth          
]

Question : \{\{\textsl{question}\}\} \\
 
Options : \\
A) \{\{\textsl{option A}\}\} \\
B) \{\{\textsl{option B}\}\} \\
... \\
 
Please reason step by step, and you should write the correct option alphabet within \textbackslash boxed\{\}.

\end{tcolorbox}
\caption{Prompt used for evaluating \model models on multiple-choice questions. We apply the prompt to the CSAT 2025, GPQA Diamond, MMLU, and MMLU-Pro benchmarks. The number of options is adjusted for each test case.}
\label{fig:mcqa_prompt_ours}
\end{figure}

\begin{figure}[ht!]
\centering
\begin{tcolorbox}[
  title=Prompt for Code Generation, 
  colframe=Black!80!White,   
  colback=gray!10,          
  coltitle=white,           
  colbacktitle=Black!80!White, 
  fonttitle=\bfseries,      
  breakable=false,          
  rounded corners,          
  boxsep=3pt,               
  width=\textwidth          
]

You will be given a question (problem specification) and will generate a correct Python program that matches the specification and passes all tests. You should first think about the step-by-step reasoning process and then provide the code.\\
\\
Question: \{\{\textsl{question}\}\}
\end{tcolorbox}
\caption{Prompt used for evaluating \model models on code generation. We apply the prompt to the LiveCodeBench Code Generation task.}
\label{fig:code_prompt_ours}
\end{figure}

\subsection{Experimental Results}

The performance comparison between \model and baseline models is conducted across four categories: mathematics, science, coding, and general knowledge. The evaluation results for the mathematics category are presented in Table~\ref{tab:math_result}, while those for the other categories are given in Table~\ref{tab:other_result}.

We find that the \model 32B model exhibits competitive performance, rivaling leading open-weight reasoning models such as DeepSeek-R1 and QwQ-32B. Notably, it outperforms the distilled versions of DeepSeek-R1, including DeepSeek-R1-Distill-Qwen-32B and DeepSeek-R1-Distill-Llama-70B.

Furthermore, the \model 7.8B model demonstrates superior performance compared to models of similar scale such as DeepSeek-R1-Distill-Qwen-7B and DeepSeek-R1-Distill-Llama-8B. It also outperforms the proprietary reasoning model: OpenAI o1-mini.

Regarding the \model 2.4B, it outperforms the DeepSeek-R1-Distill-Qwen-1.5B. Our experimental results highlight the \model models demonstrate enhanced reasoning capabilities across different model sizes.
\begin{table}[h!]
    \centering
    \small
    \begin{tabular}{wl{0.27\linewidth}|wc{0.08\linewidth}wc{0.09\linewidth}wc{0.09\linewidth}wc{0.09\linewidth}wc{0.09\linewidth}wc{0.09\linewidth}}
        \toprule
        \makecell{Model} & \makecell{MATH-500 \\ \smaller (pass@1)} & \makecell{AIME 2024 \\ \smaller (pass@1)} & \makecell{AIME 2024 \\ \smaller (cons@64)} & \makecell{AIME 2025 \\ \smaller (pass@1)} & \makecell{AIME 2025 \\ \smaller (cons@64)} & \makecell{CSAT 2025 \\ \smaller (pass@1)} \\
        \midrule
        \rowcolor[rgb]{0.9,0.9,0.9} \model 32B & \underline{95.7}~~~ & 72.1~~~ & \textbf{90.0}~~~ & 65.8 & \textbf{80.0} & \textbf{94.5} \\
        DeepSeek-R1-Distill-Qwen-32B & 94.3* & 72.6* & 83.3* & 55.2 & 73.3 & 84.1 \\
        QwQ-32B & 95.5~~~ & \underline{79.5}* & \underline{86.7}~~~ & \textbf{67.1} & \underline{76.7} & \underline{94.4} \\
        DeepSeek-R1-Distill-Llama-70B & 94.5* & 70.0* & \underline{86.7}* & 53.9 & 66.7 & 88.8 \\
        DeepSeek-R1 (671B) & \textbf{97.3}* & \textbf{79.8}* & \underline{86.7}~~~ & \underline{66.8} & \textbf{80.0} & 89.9 \\
        \midrule
        \rowcolor[rgb]{0.9,0.9,0.9} \model 7.8B & \textbf{94.8}~~~ & \textbf{70.0}~~~ & \textbf{83.3}~~~ & \textbf{59.6} & \textbf{76.7} & \textbf{89.9} \\
        DeepSeek-R1-Distill-Qwen-7B & \underline{92.8}* & 55.5* & \textbf{83.3}* & 38.5 & 56.7 & 79.7 \\
        DeepSeek-R1-Distill-Llama-8B & 89.1* & 50.4* & \underline{80.0}* & 33.6 & 53.3 & 74.1 \\
        OpenAI o1-mini & 90.0* & \underline{63.6}* & \underline{80.0}* & \underline{54.8} & \underline{66.7} & \underline{84.4} \\
        \midrule
        \rowcolor[rgb]{0.9,0.9,0.9} \model 2.4B & \textbf{92.3}~~~ & \textbf{52.5}~~~ & \textbf{76.7}~~~ & \textbf{47.9} & \textbf{73.3} & \textbf{79.2} \\
        DeepSeek-R1-Distill-Qwen-1.5B & \underline{83.9}* & \underline{28.9}* & \underline{52.7}* & \underline{23.9} & \underline{36.7} & \underline{65.6} \\
        \bottomrule
    \end{tabular}
    \vspace{2mm}    
    \caption{Comparison between \model and baseline models in the \texttt{Mathematics} category. The asterisk (*) indicates figures that have been officially reported. When calculating pass@1 score, we set $k$ in Equation~\ref{equ:pass_at_1} as 8 for MATH-500, 16 for CSAT 2025, and 64 for AIME 2024 and 2025. For AIME 2024 and 2025, we also report cons@64. For CSAT 2025, the individual scores for the elective subjects can be found in Table~\ref{tab:csat_detailed_result} within Appendix~\ref{appendix:evaluation_details}. \textbf{Bold} scores indicate the best performance, and \underline{underlined} scores mean the second best.}
    \label{tab:math_result}
\end{table}

\begin{table}[h!]
    \centering
    \small
    \begin{tabular}{wl{0.27\linewidth}|wc{0.15\linewidth}wc{0.15\linewidth}wc{0.15\linewidth}wc{0.15\linewidth}}
        \toprule
        \makecell{Model} & \makecell{GPQA Diamond \\ \smaller (pass@1)} & \makecell{LiveCodeBench \\ \smaller (pass@1)} & \makecell{MMLU \\ \smaller (accuracy)} & \makecell{MMLU-Pro \\ \smaller (accuracy)} \\
        \midrule
        \rowcolor[rgb]{0.9,0.9,0.9} EXAONE Deep 32B & \underline{66.1}~~~ & 59.5~~~ & 83.0~~~ & 74.0 \\
        DeepSeek-R1-Distill-Qwen-32B & 62.1* & 57.2* & 83.6~~~ & 77.2 \\
        QwQ-32B & 63.3~~~ & \underline{63.4}* & 87.4~~~ & 79.1 \\
        DeepSeek-R1-Distill-Llama-70B & 65.2* & 57.5* & \underline{89.1}~~~ & \underline{79.4} \\
        DeepSeek-R1 (671B) & \textbf{71.5}* & \textbf{65.9}* & \textbf{90.8}* & \textbf{84.0} \\
        \midrule
        \rowcolor[rgb]{0.9,0.9,0.9} EXAONE Deep 7.8B & \textbf{62.6}~~~ & \textbf{55.2}~~~ & \underline{75.0}~~~ & \underline{65.7} \\
        DeepSeek-R1-Distill-Qwen-7B & 49.1* & 37.6* & 57.7~~~ & 59.3 \\
        DeepSeek-R1-Distill-Llama-8B & 49.0* & 39.6* & 74.4~~~ & 60.6 \\
        OpenAI o1-mini & \underline{60.0}* & \underline{53.8}* & \textbf{85.2}* & \textbf{80.3} \\
        \midrule
        \rowcolor[rgb]{0.9,0.9,0.9} EXAONE Deep 2.4B & \textbf{54.3}~~~ & \textbf{46.6}~~~ & \textbf{65.8}~~~ & \textbf{56.4} \\
        DeepSeek-R1-Distill-Qwen-1.5B & \underline{33.8}* & \underline{16.9}* & \underline{33.8}~~~ & \underline{39.3} \\
        \bottomrule
    \end{tabular}
    \vspace{2mm}    
    \caption{Comparison between \model and baseline models in the \texttt{Science} (GPQA Diamond), \texttt{Coding} (LiveCodeBench), and \texttt{General} (MMLU and MMLU-Pro) categories. The asterisk (*) indicates figures that have been officially reported. When calculating pass@1 score, we set $k$ in Equation~\ref{equ:pass_at_1} as 6 for LiveCodeBench and 16 for GPQA Diamond. \textbf{Bold} scores indicate the best performance, and \underline{underlined} scores mean the second best.}
    \label{tab:other_result}
\end{table}

\section{Limitations}

The \model models introduced in this document are specifically fine-tuned to excel at reasoning tasks. Though their base models are instruction-fine-tuned and generally capable of following instructions, for addressing a wider range of real-world use cases, we strongly recommend utilizing the EXAONE 3.5 Instruct models \citep{lgai2004exaone3524binstruct, lgai2004exaone3578binstruct, lgai2004exaone3532binstruct}, which are optimized for practical application scenarios.

\section{Deployment}

Section~\ref{appendix:license} in the Appendix provides license information for using the \model models. Understanding the license information is essential for the legal utilization of the language model.

\section{Conclusion}

In this document, we presented three specialized reasoning models: \model 2.4B, 7.8B, and 32B. Despite the emergence of various methodologies aimed at improving reasoning capabilities, we have chosen to rely on well-established approaches such as SFT, DPO, and Online RL, achieving superior or competitive performance relative to models of comparable scale. Our results highlight the effectiveness and practicality of these proven techniques in advancing reasoning performance. At present, these models are primarily engaged in problem-solving in domains where clear answers are available, such as mathematics, science, and coding. Looking ahead, we aim to extend their capabilities into areas where answers are less clear or yet to be discovered, thereby broadening their impact and utility.

Our models are available to everyone for research purposes, and we welcome your feedback to help us improve the models. If you have any feedback or are interested in exploring commercial opportunities with our models, please reach out to \href{mailto:contact_us@lgresearch.ai}{contact\_us@lgresearch.ai}.

\newpage

\appendix

\section{Contributors}
\label{appendix:contributors}
All authors are listed in alphabetical order by last name.

\paragraph{Core Contributors}
Eunbi~Choi, Kibong~Choi, Seokhee~Hong, Junwon~Hwang, Hyojin~Jeon, Hyunjik~Jo, Joonkee~Kim, Seonghwan~Kim, Soyeon~Kim, Sunkyoung~Kim, Yireun~Kim, Yongil~Kim, Haeju~Lee, Jinsik~Lee, Kyungmin~Lee, Sangha~Park, Heuiyeen~Yeen, Hyeongu~Yun

\paragraph{Contributors}
Kyunghoon~Bae, Stanley~Jungkyu~Choi, Yemuk~Choi, Kijeong~Jeon, Gerrard~Jeongwon~Jo, Jiyeon~Jung, Hyosang~Kim, Youchul~Kim, Edward~Hwayoung~Lee, Honglak~Lee, Yongmin~Park, Sihoon~Yang, Sihyuk~Yi

\newpage

\section{Model License}
\label{appendix:license}

\textbf{EXAONE AI Model License Agreement 1.1 - NC} \\
\\
This License Agreement (“Agreement”) is entered into between you (“Licensee”) and LG Management Development Institute Co., Ltd. (“Licensor”), governing the use of the EXAONE AI Model (“Model”). By downloading, installing, copying, or using the Model, you agree to comply with and be bound by the terms of this Agreement. If you do not agree to all the terms, you must not download, install, copy, or use the Model. This Agreement constitutes a binding legal agreement between the Licensee and Licensor. \\
\\
\textbf{1. Definitions} \\
\\
\textbf{1.1 Model:} The artificial intelligence model provided by Licensor, which includes any software, algorithms, machine learning models, or related components supplied by Licensor. This definition extends to encompass all updates, enhancements, improvements, bug fixes, patches, or other modifications that may be provided by Licensor from time to time, whether automatically or manually implemented. \\
\\
\textbf{1.2 Derivatives:} Any modifications, alterations, enhancements, improvements, adaptations, or derivative works of the Model created by Licensee or any third party. This includes changes made to the Model's architecture, parameters, data processing methods, or any other aspect of the Model that results in a modification of its functionality or output.\\ 
\\
\textbf{1.3 Output:} Any data, results, content, predictions, analyses, insights, or other materials generated by the Model or Derivatives, regardless of whether they are in their original form or have been further processed or modified by the Licensee. This includes, but is not limited to, textual or numerical produced directly or indirectly through the use of the Model.\\
\\
\textbf{1.4 Licensor:} LG Management Development Institute Co., Ltd., the owner, developer, and provider of the EXAONE AI Model. The Licensor holds all rights, title, and interest in the Model and is responsible for granting licenses to use the Model under the terms specified in this Agreement. \\
\\
\textbf{1.5 Licensee:} The individual, organization, corporation, academic institution, government agency, or other entity using or intending to use the Model under the terms and conditions of this Agreement. The Licensee is responsible for ensuring compliance with the Agreement by all authorized users who access or utilize the Model on behalf of the Licensee. \\
\\
\textbf{2. License Grant} \\ 
\\
\textbf{2.1 Grant of License:} Subject to the terms and conditions outlined in this Agreement, the Licensor hereby grants the Licensee a limited, non-exclusive, non-transferable, worldwide, and revocable license to: \\
\\
a. Access, download, install, and use the Model solely for research purposes. This includes evaluation, testing, academic research, experimentation, and participation in competitions, provided that such participation is in a non-commercial context. Notwithstanding Section 3.1, the Licensee may only provide the Model or Derivatives for a competition if no commercial license is granted to the competition organizer or any third party. \\  
\\
b. Publicly disclose research results and findings derived from the use of the Model or Derivatives, including publishing papers or presentations. \\
\\
c. Modify the Model and create Derivatives based on the Model, provided that such modifications and Derivatives are used exclusively for research purposes. The Licensee may conduct experiments, perform analyses, and apply custom modifications to the Model to explore its capabilities and performance under various scenarios. If the Model is modified, the modified Model must include “EXAONE” at the beginning of its name.  \\
\\
d. Distribute the Model and Derivatives in each case with a copy of this Agreement. \\
\\
\textbf{2.2 Scope of License:} The license granted herein does not authorize the Licensee to use the Model for any purpose not explicitly permitted under this Agreement. Any use beyond the scope of this license, including any commercial application or external distribution, is strictly prohibited unless explicitly agreed upon in writing by the Licensor. \\
\newpage
\textbf{3. Restrictions}\label{textbf:Restrictions} \\
\\
\textbf{3.1 Commercial Use:} The Licensee is expressly prohibited from using the Model, Derivatives, or Output for any commercial purposes, including but not limited to, developing or deploying products, services, or applications that generate revenue, whether directly or indirectly. Any commercial exploitation of the Model or its derivatives requires a separate commercial license agreement with the Licensor. Furthermore, the Licensee shall not use the Model, Derivatives or Output to develop or improve other models. \\
\\
\textbf{3.2 Reverse Engineering:} The Licensee shall not decompile, disassemble, reverse engineer, or attempt to derive the source code, underlying ideas, algorithms, or structure of the Model, except to the extent that such activities are expressly permitted by applicable law. Any attempt to bypass or circumvent technological protection measures applied to the Model is strictly prohibited. \\
\\
\textbf{3.3 Unlawful Use:} The Licensee shall not use the Model and Derivatives for any illegal, fraudulent, or unauthorized activities, nor for any purpose that violates applicable laws or regulations. This includes but is not limited to the creation, distribution, or dissemination of malicious, deceptive, or unlawful content. \\
\\
\textbf{3.4 Ethical Use:} The Licensee shall ensure that the Model or Derivatives is used in an ethical and responsible manner, adhering to the following guidelines: \\
\\
a. The Model and Derivatives shall not be used to generate, propagate, or amplify false, misleading, or harmful information, including fake news, misinformation, or disinformation. \\
\\
b. The Model and Derivatives shall not be employed to create, distribute, or promote content that is discriminatory, harassing, defamatory, abusive, or otherwise offensive to individuals or groups based on race, gender, sexual orientation, religion, nationality, or other protected characteristics. \\
\\
c. The Model and Derivatives shall not infringe on the rights of others, including intellectual property rights, privacy rights, or any other rights recognized by law. The Licensee shall obtain all necessary permissions and consents before using the Model and Derivatives in a manner that may impact the rights of third parties. \\
\\
d. The Model and Derivatives shall not be used in a way that causes harm, whether physical, mental, emotional, or financial, to individuals, organizations, or communities. The Licensee shall take all reasonable measures to prevent misuse or abuse of the Model and Derivatives that could result in harm or injury. \\
\\
\textbf{4. Ownership} \\
\\
\textbf{4.1 Intellectual Property:} All rights, title, and interest in and to the Model, including any modifications, Derivatives, and associated documentation, are and shall remain the exclusive property of the Licensor. The Licensee acknowledges that this Agreement does not transfer any ownership rights to the Licensee. All trademarks, service marks, and logos associated with the Model are the property of the Licensor. \\
\\
\textbf{4.2 Output:} All rights, title, and interest in and to the Output generated by the Model and Derivatives whether in its original form or modified, are and shall remain the exclusive property of the Licensor. Licensee may use, modify, and distribute the Output and its derivatives for research purpose. The Licensee shall not claim ownership of the Output except as expressly provided in this Agreement. The Licensee may use the Output solely for the purposes permitted under this Agreement and shall not exploit the Output for unauthorized or commercial purposes. \\
\\
\textbf{4.3 Attribution:} In any publication or presentation of results obtained using the Model, the Licensee shall provide appropriate attribution to the Licensor, citing the Model's name and version, along with any relevant documentation or references specified by the Licensor. \\
\newpage
\textbf{5. No Warranty} \\
\\
\textbf{5.1 “As-Is” Basis:} The Model, Derivatives, and Output are provided on an “as-is” and “as-available” basis, without any warranties or representations of any kind, whether express, implied, or statutory. The Licensor disclaims all warranties, including but not limited to, implied warranties of merchantability, fitness for a particular purpose, accuracy, reliability, non-infringement, or any warranty arising from the course of dealing or usage of trade. \\
\\
\textbf{5.2 Performance and Reliability:} The Licensor does not warrant or guarantee that the Model, Derivatives or Output will meet the Licensee’s requirements, that the operation of the Model, Derivatives or Output will be uninterrupted or error-free, or that defects in the Model will be corrected. The Licensee acknowledges that the use of the Model, Derivatives or Output is at its own risk and that the Model, Derivatives or Output may contain bugs, errors, or other limitations. \\
\\
\textbf{5.3 No Endorsement:} The Licensor does not endorse, approve, or certify any results, conclusions, or recommendations derived from the use of the Model. The Licensee is solely responsible for evaluating the accuracy, reliability, and suitability of the Model for its intended purposes. \\
\\
\textbf{6. Limitation of Liability} \\
\\
\textbf{6.1 No Liability for Damages:} To the fullest extent permitted by applicable law, in no event shall the Licensor be liable for any special, incidental, indirect, consequential, exemplary, or punitive damages, including but not limited to, damages for loss of business profits, business interruption, loss of business information, loss of data, or any other pecuniary or non-pecuniary loss arising out of or in connection with the use or inability to use the Model, Derivatives or any Output, even if the Licensor has been advised of the possibility of such damages. \\
\\
\textbf{6.2 Indemnification:} The Licensee agrees to indemnify, defend, and hold harmless the Licensor, its affiliates, officers, directors, employees, and agents from and against any claims, liabilities, damages, losses, costs, or expenses (including reasonable attorneys' fees) arising out of or related to the Licensee's use of the Model, any Derivatives, or any Output, including any violation of this Agreement or applicable laws. \\
\\
\textbf{7. Termination} \\
\\
\textbf{7.1 Termination by Licensor:} The Licensor reserves the right to terminate this Agreement and revoke the Licensee’s rights to use the Model at any time, with or without cause, and without prior notice if the Licensee breaches any of the terms or conditions of this Agreement. Termination shall be effective immediately upon notice. \\
\\
\textbf{7.2 Effect of Termination:} Upon termination of this Agreement, the Licensee must immediately cease all use of the Model, Derivatives, and Output and destroy all copies of the Model, Derivatives, and Output in its possession or control, including any backup or archival copies. The Licensee shall certify in writing to the Licensor that such destruction has been completed. \\
\\
\textbf{7.3 Survival:} The provisions of this Agreement that by their nature should survive termination, including but not limited to, Sections 4 (Ownership), 5 (No Warranty), 6 (Limitation of Liability), and this Section 7 (Termination), shall continue to apply after termination. \\
\\
\textbf{8. Governing Law} \\
\\
\textbf{8.1 Governing Law:} This Agreement shall be governed by and construed in accordance with the laws of the Republic of Korea, without regard to its conflict of laws principles. \\
\\
\textbf{8.2 Arbitration:} Any disputes, controversies, or claims arising out of or relating to this Agreement, including its existence, validity, interpretation, performance, breach, or termination, shall be referred to and finally resolved by arbitration administered by the Korean Commercial Arbitration Board (KCAB) in accordance with the International Arbitration Rules of the Korean Commercial Arbitration Board in force at the time of the commencement of the arbitration. The seat of arbitration shall be Seoul, Republic of Korea. The tribunal shall consist of one arbitrator. The language of the arbitration shall be English. \\
\newpage
\textbf{9. Alterations} \\
\\
\textbf{9.1 Modifications:} The Licensor reserves the right to modify or amend this Agreement at any time, in its sole discretion. Any modifications will be effective upon posting the updated Agreement on the Licensor’s website or through other means of communication. The Licensee is responsible for reviewing the Agreement periodically for changes. Continued use of the Model after any modifications have been made constitutes acceptance of the revised Agreement. \\
\\
\textbf{9.2 Entire Agreement:} This Agreement constitutes the entire agreement between the Licensee and Licensor concerning the subject matter hereof and supersedes all prior or contemporaneous oral or written agreements, representations, or understandings. Any terms or conditions of any purchase order or other document submitted by the Licensee in connection with the Model that are in addition to, different from, or inconsistent with the terms and conditions of this Agreement are not binding on the Licensor and are void. \\
\\
By downloading, installing, or using the EXAONE AI Model, the Licensee acknowledges that it has read, understood, and agrees to be bound by the terms and conditions of this Agreement. \\

\newpage

\section{Evaluation Details}
\label{appendix:evaluation_details}

\begin{table}[h!]
    \centering
    \small
    \begin{tabular}{wl{0.27\linewidth}|wc{0.15\linewidth}wc{0.15\linewidth}wc{0.15\linewidth}}
        \toprule
        \makecell{Model} & \makecell{CSAT 2025 \\ Calculus \\ \smaller (pass@1)} & \makecell{CSAT 2025 \\ Statistics \\ \smaller (pass@1)} & \makecell{CSAT 2025 \\ Geometry \\ \smaller (pass@1)} \\
        \midrule
        \rowcolor[rgb]{0.9,0.9,0.9} EXAONE Deep 32B & \textbf{95.1} & \underline{95.0} & \textbf{93.5} \\
        QwQ-32B & \underline{94.5} & \textbf{95.5} & \underline{93.3} \\
        DeepSeek-R1 (671B) & 89.4 & 90.8 & 89.6 \\
        DeepSeek-R1-Distill-Llama-70B & 88.1 & 90.3 & 88.1 \\
        DeepSeek-R1-Distill-Qwen-32B & 84.3 & 84.4 & 83.5 \\
        \midrule
        \rowcolor[rgb]{0.9,0.9,0.9} EXAONE Deep 7.8B & \textbf{89.5} & \textbf{91.8} & \textbf{88.5} \\
        OpenAI o1-mini & \underline{82.8} & \underline{85.9} & \underline{84.6} \\
        DeepSeek-R1-Distill-Qwen-7B & 75.6 & 83.8 & 79.6 \\
        DeepSeek-R1-Distill-Llama-8B & 70.6 & 79.0 & 72.6 \\
        \midrule
        \rowcolor[rgb]{0.9,0.9,0.9} EXAONE Deep 2.4B & \textbf{74.4} & \textbf{82.4} & \textbf{80.8} \\
        DeepSeek-R1-Distill-Qwen-1.5B & \underline{64.4} & \underline{67.6} & \underline{64.6} \\
        \bottomrule
    \end{tabular}
    \vspace{2mm}    
    \caption{Comparison of \model with baseline models in the individual scores for the elective subjects at the CSAT 2025 benchmark. \textbf{Bold} scores indicate the best performance, and \underline{underlined} scores mean the second best.}
    \label{tab:csat_detailed_result}
\end{table}

\newpage
\bibliographystyle{plain} 
\bibliography{refs} 


\end{document}